\documentclass{article}

\usepackage{amssymb}
\usepackage{graphicx}

\usepackage{stackengine}

\usepackage{makeidx}
\usepackage{graphicx}
\usepackage{amsmath,amssymb} 
\usepackage{color}

\usepackage{epsfig}
\usepackage{graphicx}
\usepackage{amsmath}
\usepackage{amssymb}

\usepackage{mathtools}

\usepackage{relsize}
\usepackage{multirow}
\usepackage{bm}

\usepackage{microtype}
\usepackage{graphicx}
\usepackage{subfigure}
\usepackage{booktabs} 
\usepackage{stackengine}
\usepackage{hyperref}

\usepackage[accepted]{icml2019}

\newcommand{\icol}[1]{
  \begin{smallmatrix}#1\end{smallmatrix}%
}

\icmltitlerunning{Clustering with Assignment Based Nonlinear Transform}

\begin{document}

\twocolumn[
\icmltitle{Clustering with Jointly Learned Nonlinear Transforms Over \\ Discriminating Min-Max Similarity/Dissimilarity Assignment}




\begin{icmlauthorlist}
\icmlauthor{Dimche Kostadinov}{to}
\icmlauthor{Behrooz Razeghi}{to}
\icmlauthor{Taras Holotyak}{to}
\icmlauthor{Slava Voloshynovskiy}{to}
\end{icmlauthorlist}

\icmlaffiliation{to}{Department of Computer Science, University of Geneva, Switzerland}
\icmlcorrespondingauthor{Dimche Kostadinov}{dimche.kostadinov@unige.ch}
\icmlcorrespondingauthor{${}$}{dime.kostadinov@gmail.com}
\icmlkeywords{Machine Learning, ICML}

\vskip 0.3in
]



\printAffiliationsAndNotice{}  

\begin{abstract}

This paper presents a novel clustering concept that is based on jointly learned nonlinear transforms (NTs) with priors on the information loss and the discrimination. We introduce a clustering principle that is based on evaluation of a parametric min-max measure for the discriminative prior. The decomposition of the prior measure allows to break down the assignment into two steps. In the first step, we apply NTs to a data point in order to produce candidate NT representations. In the second step, we preform the actual assignment by evaluating the parametric measure over the candidate NT representations. Numerical experiments on image clustering task validate the potential of the proposed approach. The evaluation shows advantages in comparison to the state-of-the-art clustering methods.

\end{abstract}

\section{Introduction}
\label{sec:introduction}

Clustering is one of the most important unsupervised learning task in the areas of signal processing, machine learning, computer vision and artificial intelligence that has been extensively studied for decades. 
Commonly, 
the 
data clustering algorithms 
\cite{Cover:2006:EIT:1146355}, 
\cite{Hoyer04non-negativematrix}, 
\cite{DBLP:conf:accv12:GuoJD}, 
\cite{DBLP:journals/pami/JiangLD13}, 
\cite{DBLP:eccv14:CaiZZFW:1}, 
\cite{DBLP:conf:icip:ShekharPC14}, 
\cite{xu2005maximum}, 
\cite{bach2008diffrac}  
and 
\cite{krause2010discriminative} 
address the problem of identification and description of the underlining clusters that explain the data. 

Among the various 
types of clustering algorithms,  the k-means and matrix decomposition based methods are one of the most popular and practically useful approaches. 
Given a data set, in the most common case, the objective of a clustering algorithm is to minimize the inter-cluster cost, \textit{i.e.}, the measured similarity between the  data cluster and the data points under that cluster  and maximize the intra-cluster cost, \textit{i.e.}, the measured similarity between the data cluster and the data points that do not belong to that cluster. 
A data factorization/decomposition model \cite{Cover:2006:EIT:1146355} \cite{Hoyer04non-negativematrix} \cite{krause2010discriminative} \cite{vidal2011subspace} with constraints summarizes a  general problem formulation that also subsumes 
the previously explained basic case. We express it in the following: 
\vspace{-.15in}
\begin{equation}
\begin{aligned}
\!\!\!\!\!\!\!\{\hat{\bf Y}, \hat{\bf C}, \hat{\boldsymbol{\theta}}\} =\arg \min_{{\bf Y}, {\bf C}, \boldsymbol{\theta}} \sum_{i=1}^{CK}\left[ g( {\bf x}_i,{\bf C}{\bf y}_i )   +  \lambda_0 f_s (  {\bf y}_i )\right.  + \\
\!\!\!\!\!\!\! \left. \lambda_1f_c( {\bf y}_i, \boldsymbol{\theta} ) \right] +  \lambda_2f_d({\bf C}),
\end{aligned}
\label{intro:syntesys:model}
\end{equation}
where ${\bf C}=[{\bf c}_1,...,{\bf c}_C] \in \Re^{N \times C}$ are the clusters, ${\bf x}_i \in \Re^{N}$ is the $i$-th data point, ${\bf y}_i \in \Re^{C}$ is its data representation over the clusters, $\boldsymbol{\theta} $ 
are parameters responsible for a tasks specific functionality,  
$g(.,.)$ is the similarity measure between the data point and the representation over the clusters, $f_c( ., .)$ and $f_s (  . )$ are the task specific and sparsity penalty functions, respectively, $f_d(.)$ is penalty on the cluster properties and $\{\lambda_0, \lambda_1, \lambda_2 \}$ are Lagrangian parameters. 
The cluster assignment in 
\eqref{intro:syntesys:model} is based on the \textit{synthesis model}
\cite{Aharon:2006:SAD:2197945.2201437}, \cite{RubinsteinR:ADL}, where usually 
${\bf x}_i$ is reconstructed and represented by a sparse linear combination ${\bf y}_i$ over the clusters ${\bf C}$ as ${\bf x}_i \simeq {\bf C}{\bf y}_i$. 
In essence, the crucial element behind this clustering principle 
is the used measure $g(.,.)$ for similarity 
as well as the penalty functions $f_c( ., .)$, $f_s (  . )$ and $f_d ( . )$  which have significant role 
in the cluster vectors estimation and impact the resulting cluster assignment. 
Due to the used model, the solution to \eqref{intro:syntesys:model} might not only have high computational complexity, 
but also there might be difficulties in modeling and imposing constraints ($f_c( ., .)$, $f_s (  . )$ and $f_d ( . )$ in \eqref{intro:syntesys:model}) that are requered in order to preserve specific data properties, like structured sparsity \cite{Hoyer04non-negativematrix}, pairwise constraints \cite{DBLP:conf:icip:ShekharPC14}, data subspace 
\cite{elhamifar2009sparse},
 \cite{vidal2011subspace}, 
 \cite{ lu2012robust}, graph structure and manifold curvature  \cite{krause2010discriminative} and 
\cite{daitch2009fitting}. 

On the other hand, beside the synthesis model, 
in the area of signal processing, the other two commonly used models are 
the \textit{analysis model} \cite{RubinsteinPE13} 
and the \textit{sparsyfying transform model} 
\cite{RubinsteinE14} and \cite{DBLP:conf/icassp/RavishankarB14}. In the transform model, ${\bf y}_{i}$ is a nonlinear transform (NT) representation that is estimated using a linear mapping ${\bf A}{\bf x}_{i}$, with map ${\bf A} \in \Re^{M \times N}$, which then is followed by an element-wise nonlinearity and it represents a solution to a \textit{constrained projection problem}. Under this model, the computational complexity for estimating the representation ${\bf y}_i$ is low, but, so far, 
 was not addressed, nor considered
 as basis 
 for clustering or learning discriminative NT representations.  
In addition, in spite the fact that an NT model offers a high degree of freedom for modeling a wide class of constraints\footnote{Many nonlinearities, i.e., ReLu, $p$-norms, elastic net-like, $\frac{\ell_1}{\ell_2}$-norm ratio, binary encoding, ternary encoding, etc., can be expressed and modeled by a nonlinear transform.}, 
robust assignment cost under NT model that is based on a parametric measure which jointly 
takes into account not only similarity, but also dissimilarity contribution, was not studied nether explored.

\subsection{Nonlinear Transform Model, Assignment Principle and Learning Strategy Outline}

In this paper, 
we introduce  
an assignment based nonlinear transform model for clustering.

\textbf{Assignment Based NT Model} 
We addresses the problem of estimating the parameters that model the probability $p({\bf y}_{i}| {\bf x}_{i}, {\bf A})=\int_{\boldsymbol{\theta}}p({\bf y}_{i}, \boldsymbol{\theta}| {\bf x}_{i}, {\bf A})d\boldsymbol{\theta}$ of assigning a cluster and a nonlinear transform representation ${\bf y}_i \in \Re^M$ for input data ${\bf x}_i \in \Re^N$ by using the parameters $\boldsymbol{\theta}$ and ${\bf A}\in \Re^{M \times N}$.  

We motivate the use 
of ${\bf A}$, 
in order to extract useful 
data properties.  
If $M = N$, a suitable prior allows us to model a metric (linear map) to achieve invariance \cite{Li:2016:TIS}. In this paper, we model overcomplete ${\bf A}$ with $M>N$ using the prior $p({\bf A})$. Essentially, with our prior, we introduce redundancy in a constrained way, while we approximatively preserve the properties of the original data in order to pronounce discrimination among the assigned  
NT representations ${\bf y}_i$. 
Nonetheless, we note that even in the case of ${\bf A}={\bf I}$,  we can use our model, which reduces to a nonlinear assignment in the original space of ${\bf x}_i$, \textit{i.e.}, $p({\bf y}_{i}|{\bf x}_{i})=\int_{\boldsymbol{\theta}}p({\bf y}_{i}, \boldsymbol{\theta}| {\bf x}_{i})d\boldsymbol{\theta}$. 
Our assignment measure is nonlinear. 
In general, not all nonlinear function can highlight relevant data properties that are related to discrimination. 
We consider a piece-wise linear nonlinarity. In order to address the robustness 
in the assignment, we explicitly model parameters $\boldsymbol{\theta}=\{\boldsymbol{\theta}_1, \boldsymbol{\theta}_2 \}$ related to both similarity and dissimilarity contribution.  
Their role is the discrimination functionality that we address by using a composite 
min-max assignment measure. 
In its evaluation,  due to the decomposition of the min-max assignment measure, the pair $\{\boldsymbol{\tau}_{c_1}, \boldsymbol{\nu}_{c_2} \}$ from $\{\boldsymbol{\theta}_1, \boldsymbol{\theta}_2 \}$ $=\{ \{ \boldsymbol{\tau}_1,...,\boldsymbol{\tau}_{C_d} \}, $ $ \{ \boldsymbol{\nu}_1,...,\boldsymbol{\nu}_{C_s} \} \} \in \Re^{M \times (C_d+C_s)}$
has additional interpretation. It is viewed as NT specific parameter, 
which is used to produce the respective candidate NT representation.  

\textbf{Cluster and NT Representation Assignment} 
During 
cluster assignment, instead of description by clusters, we rely on candidate NTs. 
We estimate a single candidate NT representation 
as a solution to a \textit{direct}, \textit{constrained projection problem}. 
To attain unique and distinctive patterns in the resulting candidate NT representation, we parameterize the corresponding candidate NT with shared 
${\bf A}$ and distinct $\{\boldsymbol{\tau}_{c_1}, \boldsymbol{\nu}_{c_2} \}$ that 
are used for the 
element-wise nonlinearity. 

When \eqref{intro:syntesys:model} is used, ${\bf x}_i$ has only one representation 
that usually under certain similarity score is related to the likelihood of the assignment w.r.t. clusters. 
On the other hand, in our model, 
we apply a number of candidate NTs on a data point ${\bf x}_i$, 
which results in a number of candidate NT representations.  
Afterwords, our assignment is based on evaluating a min-max similarity/dissimilarity score using all  
of the candidate NT representations 
and their corresponding 
parameters $\{\boldsymbol{\tau}_{c_1}, \boldsymbol{\nu}_{c_2} \}$.  
Nonetheless, based on the same assignment score, we describe ${\bf x}_i$  
by only one 
candidate NT representation. 
In fact, in this way, we simultaneously assign both the cluster index and the NT representation ${\bf y}_i$ based on the evaluation of the min-max measure. 




\textbf{Learning Strategy} In order to estimate the parameters of our model, we consider 
$p({\bf Y}, {\bf A}| {\bf X})=p({\bf Y}| {\bf X}, {\bf A})p({\bf A}|{\bf X})=\prod_{i=1}^{CK}\int_{\boldsymbol{\theta}}p({\bf y}_i, \boldsymbol{\theta}| {\bf x}_i, {\bf A})d\boldsymbol{\theta} p({\bf A}|{\bf x}_i)$. 
Its maximization over ${\bf Y},\boldsymbol{\theta}$ and ${\bf A}$ is difficult. 
We address 
a 
point-wise approximation to the marginal $\int_{\boldsymbol{\theta}}p({\bf y}_i, \boldsymbol{\theta}| {\bf x}_i, {\bf A})d\boldsymbol{\theta}$, which allows us to derive an efficient learning algorithm.  

Compared with the factorization based clustering methods \eqref{intro:syntesys:model}, the fundamental difference of our approach is the used model.
The factorization/decomposition model addresses the joint \textit{data reconstruction} and cluster estimation with constraints by solving \textit{inverse problem}, w.r.t. the model ${\bf x}_i = {\bf C}{\bf y}_i+{\bf e}_i$, where ${\bf e}_i \in \Re^{N}$ is the error vector. In our assignment based nonlinear transform model, we address joint learning of 
\textit{data projections} (NTs) 
with information loss and discriminative priors, 
by solving \textit{direct}, \textit{constrained projection problems}, based on the candidate models in the form ${\bf A}{\bf x}_i = {\bf y}_i+{\bf z}_i$, where ${\bf z}_i \in \Re^{M}$ is the NT error vector.

\subsection{Contributions}
In the following, we outline our contributions. 

(i) We introduce novel cluster assignment principle 
that is centered on two elements: (1) joint modeling and learning of 
\textit{nonlinear transforms} (NTs) with priors 
and (2) \textit{cluster and NT representation assignment} based on 
a 
min-max 
score. 
To the best of our knowledge, our novel 
discriminative 
assignment principle is first of this kind that:
\begin{enumerate}
\vspace{-0.1in}
\item[(a)] Introduces a clustering concept that is based on modeling a direct problem
\vspace{-0.1in}
\item[(b)] Addresses a trade-off between robustness in the cluster assignment and the NT representaion compactness by allowing reduction or extension of the NT dimensionality while increasing or decreasing the number of the discrimination parameters $\boldsymbol{\theta}$
\vspace{-0.1in}
\item[(c)] Offers 
cluster assignment over a wide class of similarity score functions including a min-max while enabling efficient estimation of the NT representation
\vspace{-0.1in}
\item[(d)] Allows a rejection option and cluster grouping over continues, discontinues and overlapping regions in the transform domain.
\end{enumerate}
\vspace{-0.1in}


(ii) We propose an efficient learning strategy in order to estimate the parameters of the NTs. We implement it by an iterative alternating algorithm with three steps. At each step we give an exact and approximate closed form solution. 

(iii) We present numerical experiments that validate our model and learning principle on several 
image data sets. Our preliminary results on an image clustering task demonstrate advantages in comparison to the state-of-the-art methods, w.r.t. the computational efficiency in training and test time and the used clustering performance measures. 


\section{Related Work}

In the following, we describe the related prior work. 

\textbf{K-means, Matrix Factorization Models and Dictionary Learning} 
Factor analysis \cite{child2006essentials} and matrix factorization \cite{Hoyer04non-negativematrix} relay on decomposition on hidden features without or with constraints. One special case with only a constraint on the sparsity of the hidden representation, which is considered as a "hard" assignment is the basic k-means \cite{Cover:2006:EIT:1146355} algorithm. When discrimination constraints are present, they act as regularization, which were mainly defined using labels in the discriminative dictionary learning methods 
\cite{DBLP:journals/pami/JiangLD13}, \cite{DBLP:eccv14:CaiZZFW:1} and \cite{DBLP:conf:icip:ShekharPC14}. 

\textbf{Kernel, Subspace and Manifold Based Clustering} 
Intended to capture the nonlinear structure of the data with outliers and noise, the kernel k-means algorithms \cite{Dhillon:2004:KKS} and \cite{Chitta:2011:AKK} have been proposed. 
Also, many subspace clustering methods
were proposed \cite{vidal2011subspace}, \cite{ ma2008estimation}, \cite{ma2007segmentation}, \cite{ lu2012robust}, \cite{elhamifar2009sparse} and \cite{bradley2000k}. Commonly they consist of (i) 
subspace learning via matrix factorization and (ii) grouping of the data
into clusters in the learned subspace. 
Some authors \cite{daitch2009fitting} even include a graph regularization into the subspace clustering. 

\textbf{Discriminative Clustering} 
In \cite{xu2005maximum} clustering with maximum margin constraints was proposed. The authors in \cite{bach2008diffrac} proposed linear clustering based on a linear discriminative cost function with convex relaxation. In \cite{krause2010discriminative}
regularized information maximization was proposed and simultaneous clustering and classifier training was preformed. 
The above methods rely on kernels and account high computational complexity. 
\textbf{Self-Supervision, Self-Organization and Auto-Encoders}  
In self-supervised learning \cite{doersch2015unsupervised}, \cite{pathak2016context} the input data determine the labels. In self-organization \cite{kohonen1982self}, \cite{vesanto2000clustering} 
a neighborhood function is 
used to preserve the topological properties of the input space. Both of the approaches leverage implicit discrimination using the data. 
The single layer auto-encoder \cite{Baldi:2011AE} and its denoising extension \cite{VincentLLBM10} consider robustness to noise and reconstruction. While the idea is to encode and decode the data using a reconstruction loss, an explicit constraint that enforces discrimination is not addressed.

\section{Assignment Based NT With Priors} 

In the following, we introduce our deterministic model and show how $\boldsymbol{\tau}_{c_1}$ and $\boldsymbol{\nu}_{c_2}$ are used in the NTs to produce the candidate representations and how we perform cluster and NT representation assignment. 
Given ${\bf A}$ and ${\boldsymbol{\theta}}$, we model an assignment over 
$C_dC_s$ candidate nonlinear transforms:
\begin{align}
\mathcal{T}_T & =\{ \mathcal{T}_{\mathcal{P}_{1,1}}, ..., \mathcal{T}_{\mathcal{P}_{C_d,C_s}} \},
\end{align}
which are defined by the corresponding set of parameters: 
\begin{align}
\mathcal{P}_T  = &\{{\mathcal{P}_{1,1}}, ...,{\mathcal{P}_{C_d,C_s}} \}, \text{  where:} \nonumber \\
\mathcal{P}_{c_1,c_2}  = & \{ {{\bf A}} \in \Re^{M \times N}, \boldsymbol{\tau}_{c_1} \in \Re^{M}, \boldsymbol{\nu}_{c_2} \in \Re^{M} \}, \\ 
\label{intro:joint:transform.0}
\{ c_1, c_2 \}  \in &\mathcal{C}_d \times \mathcal{C}_s, \mathcal{C}_d \in \{1,...,C_d  \}, \mathcal{C}_s \in \{1,...,C_s  \}. \nonumber
\end{align}
All of the candidate nonlinear transforms $\mathcal{T}_{\mathcal{P}_{c_1,c_2}}$ in the set $\mathcal{T}_T$ share the linear map ${\bf A}$ and have distinct 
$\boldsymbol{\tau}_{c_1}$ and $\boldsymbol{\nu}_{c_2}$. 
A single $\mathcal{T}_{\mathcal{P}_{c_1,c_2}}$ from the set $\mathcal{T}_{T}$ 
is 
indexed using the index pair $\{ c_1, c_2 \}$ or using the single index computed as $c=c_2+(c_1-1)C_s$, $c_1 \in \mathcal{C}_d$ and $c_2 \in \mathcal{C}_s$. 

A compact description of our assignment model that takes into account $C_dC_s$ candidate NT representations while evaluating a parametric discrimination score is the following:
\begin{equation}
\begin{aligned}
f_{\boldsymbol{\theta}}: {\bf y}_i=&{\bf y}|_{\{j_1(i),j_2(i)\}}, 
\end{aligned}
\label{intro:joint:model:2.0.1.1}
\end{equation} 
\begin{equation}
\begin{aligned}
\! \! \! \! \! \! \! \! \! \{\overbrace{\hat{c}_1}^{j_1(i)},\overbrace{\hat{c}_2}^{j_2(i)}\} \! \simeq \!  \arg \min_{c_1 \in \mathcal{C}_d} \! & \! \left[ \frac{\varrho({\bf y}|_{\{{c_1}, {c_2} \}}, \boldsymbol{\tau}_{c_1})}{\max_{c_2\in \mathcal{C}_s}\varrho({\bf y}|_{\{{c_1}, {c_2} \}}, \boldsymbol{\nu}_{c_2})} \right.+\\
&\left. \varsigma({\bf y}|_{\{ {c_1}, {c_2} \}}, \boldsymbol{\tau}_{c_1}) )\right],
\end{aligned}
\label{intro:joint:model:2.0.1.0}
\end{equation} 
where $\varrho(.)$ and $\varsigma(.)$ are measures, which will be explained in details in the following subsection. A single candidate \textit{nonlinear transform} model 
is defined as follows:
\begin{equation}
\begin{aligned}
{\bf A}{\bf x}_i=&{\bf y}|_{\{c_1, c_2\}}+{\bf z}|_{\{c_1, c_2\}},\text{ where}: \\
{\bf y}|_{\{c_1, c_2\}}=&\mathcal{T}_{\mathcal{P}_{c_1,c_2} }( {\bf x}_{i} ),
\end{aligned}
\label{intro:joint:model:2.0.1}
\end{equation} 
and $\mathcal{T}_{\mathcal{P}_{c_1,c_2}}({\bf x}_{i}): \Re^{N} \rightarrow \Re^{M}$ is the parametric candidate NT that produces the candidate NT representation ${\bf y}|_{\{ c_1,c_2\}}$, by using the set of parameters $\mathcal{P}_{c_1,c_2}=\{ {\bf A}, \boldsymbol{\tau}_{c_1}, \boldsymbol{\nu}_{c_2}  \}$, while ${\bf z}|_{\{c_1, c_2\}} \in \Re^{M}$ is the NT error vector.

\subsection{The Probabilistic Assignment Based NT Model}

In probability, we use the following model: 
\begin{equation}
\begin{aligned}
p({\bf y}_i|{\bf x}_i,{\bf A})=\int_{\boldsymbol{\theta}}p({\bf y}_{i}, \boldsymbol{\theta}| {\bf x}_{i}, {\bf A})d\boldsymbol{\theta}.
\end{aligned}
\label{intro:joint:model.consitional}
\end{equation}
Furthermore, we use the Bayes' rule, disregard the prior $p({\bf x}_i|{\bf A})$ and focus on the proportional relation, 
\textit{i.e.}:
\begin{equation}
\begin{aligned}
p({\bf y}_{i}, \boldsymbol{\theta}|{\bf x}_{i}, {\bf A}) \propto p({\bf x}_{i}|{\bf y}_{i}, \boldsymbol{\theta},  {\bf A})p({\bf y}_{i}, \boldsymbol{\theta}|{\bf A}).
\end{aligned}
\label{intro:joint:model.0.0.1}
\end{equation} 
In our model, 
the probability $p({\bf x}_{i}|{\bf y}_{i}, \boldsymbol{\theta},  {\bf A})$ takes into account the  NT error and the discrimination parameter adjustment error. While $p({\bf y}_{i}, \boldsymbol{\theta}|{\bf A})$ is the parametric discriminative prior, which 
we simplify it by the following assumption: 
\begin{equation}
\begin{aligned}
p(\boldsymbol{\theta}, {\bf y}_{i}|{\bf A})=p(\boldsymbol{\theta}, {\bf y}_{i}), 
\end{aligned}
\label{intro:joint:model:1}
\end{equation} 
where we disregard the dependences on ${\bf A}$. We denote the discrimination related parameters as: 
\begin{equation}
\begin{aligned}
\! \! \! \! \! \! \boldsymbol{\theta}=&\{\boldsymbol{\theta}_1,\boldsymbol{\theta}_2 \},\\
\text{ dissimilarity: } \boldsymbol{\theta}_1=&\{\boldsymbol{\tau}_1,...,\boldsymbol{\tau}_{C_d} \}, \\
\text{ similarity: }\boldsymbol{\theta}_2=&\{\boldsymbol{\nu}_1,...,\boldsymbol{\nu}_{C_s} \},
\end{aligned}
\label{intro:joint:model:theta:parameters.0}
\end{equation} 
which are used in our min-max assignment over dissimilarity and similarity contributions w.r.t. all pairs $\boldsymbol{\tau}_{c_1}$ and $\boldsymbol{\nu}_{c_2}$.

\subsection{Assignment and Adjustment Modeling} 

We model the assignment based nonlinear transform together with the discrimination parameters 
using the r.h.s. of \eqref{intro:joint:model.0.0.1}. 

Using two measures, we define $p({\bf x}_{i}| {\bf y}_{i}, \boldsymbol{ \theta}, {\bf A})$ as follows:
\begin{equation}
\begin{aligned}
p({\bf x}_{i}| &{\bf y}_{i}, \boldsymbol{ \theta}, {\bf A}) \propto \\
& \exp \left[ -\frac{1}{\beta_0}u_r({\bf A}{\bf x}_i,{\bf y}_i)-\frac{1}{\beta_a}u_a({\bf A}{\bf x}_i, \boldsymbol{\theta}) \right],
\end{aligned}
\label{intro:joint:model:2}
\end{equation} 
where $u_r({\bf A}{\bf x}_i,{\bf y}_i)$ and $u_a({\bf A}{\bf x}_i, \boldsymbol{\theta})$ take into account the  NT and the discrimination parameter adjustment errors, respectively, while $\beta_0$ and $\beta_a$ are scaling parameters.

{\textbf{NT Error}} Note that in \eqref{intro:joint:model:2.0.1.1},  ${\bf y}_{i}$ is evaluated using an assignment over the candidate NT representations ${\bf y}|_{\{c_1,c_2 \}}$ that result from applying the nonlinear transform $\mathcal{T}_{\mathcal{P}_{c_1,c_2} }( {\bf x}_{i} )$ on ${\bf A}{\bf x}_i$. Therefore,  
we can say that the term ${\bf z}|_{\{j_1(i),j_2(i)\} }={\bf A}{\bf x}_{i} -{\bf y}|_{\{j_1(i),j_2(i)\} }$, \textit{i.e.}, ${\bf z}_{i}={\bf A}{\bf x}_{i}-{\bf y}_{i}$ is the \textit{nonlinear transform error} vector that represents the deviation of ${\bf A}{\bf x}_{i}$ from the targeted transform representation ${\bf y}_{i}$. 
In the simplest form, we assume ${\bf z}_{i}$ to be Gaussian distributed and we model: 
\begin{equation}
\begin{aligned}
u_r({\bf A}{\bf x}_i,{\bf y}_i)=\Vert {\bf A}{\bf x}_{i}-{\bf y}_{i}\Vert_2^2.
\end{aligned}
\label{eq:Model:ajust:parameters:model:0}
\end{equation} 

{ \textbf{NT Parameters Adjustment}} We assume that the adjustment of any of the NT discrimination parameters is w.r.t. the following  measure: 
\begin{equation}
\begin{aligned}
u_a( {\bf A}{\bf x}_{i},  \boldsymbol{\theta})=\Vert {\bf A}{\bf x}_{i}-\boldsymbol{\tau}_{j_1(i)}-\boldsymbol{\nu}_{j_2(i)} \Vert_2^2,
\end{aligned}
\label{eq:Model:ajust:parameters:model}
\end{equation}
where $j_1(i)$ and $j_2(i)$ denote the indexes of the corresponding $\boldsymbol{\tau}_{j_1(i)}$ and $\boldsymbol{\nu}_{j_2(i)}$ that are related to the assigned ${\bf y}|_{ \{j_1(i),j_2(i) \} }$ from the set of $C_dC_s$ candidate representations ${\bf y}|_{ \{c_1,c_2\} }$ under model \eqref{intro:joint:model:2.0.1.1} with the criteria \eqref{intro:joint:model:2.0.1.0}.

With \eqref{eq:Model:ajust:parameters:model} and the discrimination parameter prior (that we will describe in the next subsection), we assume that the linear transform representation ${\bf A}{\bf x}_{i}$ decomposes into two distinct components, which respectively are related to the dissimilarity and similarity parameters $\boldsymbol{\tau}_{j_1(i)}$ and $\boldsymbol{\nu}_{j_2(i)}$. 
The prior \eqref{eq:Model:ajust:parameters:model}  is crucial for a proper adjustment between the linear mapping and the pairs $\{\boldsymbol{\tau}_{j_1(i)}, \boldsymbol{\nu}_{j_2(i)} \}$ as well as for enabling the candidate NTs to discriminate in the transform domain based on their respective pair $\{\boldsymbol{\tau}_{c_1}, \boldsymbol{\nu}_{c_2} \}$. 

\subsection{Priors Modeling} 
\label{prir:measure}
A prior $p({\bf A})$ is used to allow adequate regularization of the coherence and conditioning  
 on the transform matrix ${\bf A}$, whereas the joint modeling of the $C_sC_d$ NTs is enabled by using the prior $p(\boldsymbol{ \theta},{\bf y}_{i})$.

{\flushleft \textbf{Minimum Information Loss Prior }} 
By the term "minimum information loss" we mean that the linear map ${\bf A}$ approximatively preserves the data properties  
in the transform space. In order to simplify, we assume that $p({\bf A}|{\bf X})=p({\bf A})$ and define our prior $p({\bf A})$ as $p({\bf A}) \propto \exp (-f_d({\bf A}) )$, where $f_d({\bf A})=(\frac{1}{\beta_3}\Vert {\bf A} \Vert_{F}^2+\frac{1}{\beta_4}\Vert {\bf A}{\bf A}^T-{\bf I} \Vert_F^2-\frac{1}{\beta_5}\log \vert \det {\bf A}^T{\bf A} \vert )$ \cite{DBLP:conf/icassp/RavishankarB14} and \cite{Kostadinov2018:EUVIP}. 
Under this prior measure, we essentially relate our notion of "information loss" by constraining the conditioning and the expected coherence of ${\bf A}$. 


\label{Priors:Modeling}
\vspace{-0.07in}
{\flushleft \textbf{Discrimination Prior}} 
We model a discrimination prior as: 
\begin{equation}
\begin{aligned}
p(&\boldsymbol{\theta},{\bf y}_{i}) \propto \\
&\exp \left[ -\frac{1}{\beta_d}f_c(\boldsymbol{\theta}, {\bf y}_{i})- \frac{1}{\beta_E}u_p(\boldsymbol{\theta})-\frac{1}{\beta_1}\Vert {\bf y}_{i}\Vert_1 \right],  
\end{aligned}
\label{eq:Model:discriminate:parameters:model}
\end{equation}
where 
$f_c(\boldsymbol{\theta}, {\bf y}_{i})$ and $u_p(\boldsymbol{\theta})$ are NT representation and NT parameters related measures, respectively, that have discrimination role,  
while $\Vert {\bf y}_{i}\Vert_1$ is our 
sparsity 
measure 
and $\beta_1, \beta_d$ and $\beta_E$ are scaling parameters.  

{$-$ \underline{\textit{Compositional Min-max Discrimination Measure}}}  To define $f_c(\boldsymbol{\theta}, {\bf y}_{i})$ we assume that: 
\begin{enumerate}
\vspace{-0.1in}
\item[(i)] The relation between $\boldsymbol{\theta}$ and ${\bf y}_{i}$ is determined on the \textit{ vector support intersection} between ${\bf y}_{i}$, $ \boldsymbol{\tau}_{c_1}$ and $ \boldsymbol{\nu}_{c_2}$
\vspace{-0.1in}
\item[(ii)] The min-max description is  decomposable w.r.t. $\boldsymbol{\tau}_{c_1}$ and $\boldsymbol{\nu}_{c_2}$ 
\vspace{-0.1in}
\item[(iii)] The support intersection relation is specified based on  
two measures defined on the support intersection. 
\end{enumerate}


We define the two measures $\varrho$ and $\varsigma$ as $\varrho({\bf y}_{i},{\bf y}_{j})= \Vert {\bf y}_{i}^- \odot {\bf y}_{j}^- \Vert_1+\Vert {\bf y}_{i}^+ \odot {\bf y}_{j}^+ \Vert_1$ and $\varsigma( {\bf y}_{i}, {\bf y}_{j} ) = \Vert {\bf y}_{i} \odot {\bf y}_{j} \Vert_2^2$,  
where 
${\bf y}_{i}={\bf y}_{i}^+-{\bf y}_{i}^-$,${\bf y}_{j}={\bf y}_{j}^+-{\bf y}_{j}^-$, ${\bf y}_{i}^+=$ $\max({\bf y}_{i}, {\bf 0})$ and ${\bf y}_{i}^-=\max(-{\bf y}_{i},$ $ {\bf 0})$. 
The measure $\varrho({\bf y}_{i},$ $ {\bf y}_{j})$ represents our 
similarity score\footnote{When ${\bf y}_{i}^T{\bf y}_{j}$ is considered, $\varrho({\bf y}_{i}, {\bf y}_{j})$ captures contribution for the similarity, whereas $\Vert {\bf y}_{i}^+\odot {\bf y}_{j}^-\Vert_1+\Vert {\bf y}_{i}^-\odot {\bf y}_{j}^+\Vert_1$ captures  contribution for the dissimilarity between the vectors ${\bf y}_{i}$ and ${\bf y}_{j}$.}. 
On the other hand, 
$\varsigma$ measures only the strength on the support intersection.  
We use these measure to allow 
a discrimination constraint  
without any explicit assumption about the 
space/manifold in the transform domain. 

Based on the above assumptions (i), (ii) and (iii), $f_c({\bf y}_{i},  \boldsymbol{\theta})$ is defined as follows:
\vspace{-0.0in}
\begin{align}
\! \! \! \! \! \! \! \! \! \! f_c({\bf y}_{i},  \boldsymbol{\theta}) \! = \! \min_{c_1 \in \mathcal{C}_d }\left[ \frac{\varrho({\bf y}_{i},  \boldsymbol{\tau}_{c_1})}{\max_{c_2 \in \mathcal{C}_s } \varrho({\bf y}_{i}, \boldsymbol{\nu}_{c_2})} +\varsigma({\bf y}_{i},  \boldsymbol{\tau}_{c_1}  ) \right]. \label{prod:min:max}
\end{align}
The measure \eqref{prod:min:max} 
ensures that ${\bf y}_{i}$ in the transform domain will be located at the point where:

\begin{enumerate}
\vspace{-0.1in}
\item[(a)] The similarity contribution w.r.t. $\boldsymbol{\tau}_{c_1}$ is the smallest measured w.r.t. $\varrho(.)$
\vspace{-0.1in}
\item[(b)] The strength of the support intersection w.r.t. $\boldsymbol{\tau}_{c_1}$ is the smallest measured w.r.t.  $\varsigma(.)$
\vspace{-0.1in}
\item[(c)] The similarity contribution w.r.t. $\boldsymbol{\nu}_{c_2}$ is the largest  measured w.r.t. $\varrho(.)$, \textit{i.e.}, smallest w.r.t. $\frac{1}{\varrho(.)}$.
\end{enumerate}



\vspace{-0.1in}
{$-$\underline{
\textit{Discrimination Parameters Prior Measure}} } 
The measure $u_p(\boldsymbol{\theta})$ is defined as:
\vspace{-0.07in}
\begin{equation}
\begin{aligned}
u_p(\boldsymbol{\theta})= &  \sum_{c_1=1}^{C_d}f_c(\boldsymbol{\tau}_{c_1}, \boldsymbol{\theta}_{\setminus c_1 } )+\sum_{c_2=1}^{C_s}f_c(\boldsymbol{\nu}_{c_2}, \boldsymbol{\theta}_{\setminus c_2} ), \nonumber
\text{ where}:
\end{aligned}
\label{SP:COVER:measure}
\end{equation}
\vspace{ -0.1in}
\begin{equation}
\begin{aligned}
&\boldsymbol{\theta}=\{ \boldsymbol{\theta}_{1}, \boldsymbol{\theta}_{2} \}, \text{ while: }\\
&\boldsymbol{\theta}_{\setminus c_1 }=\{  \{ \boldsymbol{\tau}_{1},..., \boldsymbol{\tau}_{c_1-1},\boldsymbol{\tau}_{c_1+1},..., \boldsymbol{\tau}_{C_d} \}, \boldsymbol{\theta}_{2} \}, \\
&\boldsymbol{\theta}_{\setminus c_2 }=\{ \boldsymbol{\theta}_{1}, \{ \boldsymbol{\nu}_{1},...,\boldsymbol{\nu}_{c_2-1},\boldsymbol{\nu}_{c_2+1},..., \boldsymbol{\nu}_{C_s} \} \}. 
\label{SP:COVER:measure>plus.theta}
\end{aligned}
\end{equation}
The advantage of using \eqref{SP:COVER:measure} is that: (i) it allows non-uniform cover of the transform space in arbitrarily coarse or dense way, (ii) it gives a possibility to represents a wide range of transform space regions, including non-continues, continues and overlapping regions and (iii) at the same time it enables $\boldsymbol{\theta}$ to be described and concentrated on the most important part of the transform space related to discrimination.

\section{Problem Formulation and Learning Algorithm}

Minimizing the 
exact negative logarithm of our learning model $p({\bf Y},{\bf A} | {\bf X} )=p({\bf Y} | {\bf X}, {\bf A} )p({\bf A}|{\bf X})= \prod_{i=1}^{CK}\left[ \int_{\boldsymbol{\theta} }  p({\bf y}_{i}, \boldsymbol{\theta} | {\bf x}_{i}, {\bf A} )d\boldsymbol{\theta}\right] p({\bf A}|{\bf x}_i)$ 
over ${\bf Y}, \boldsymbol{\theta}$ and ${\bf A}$ is difficult since we have to integrate in order to compute the marginal and the partitioning function of the prior \eqref{eq:Model:discriminate:parameters:model}. 

\subsection{Problem Formulation}

Instead of minimizing the exact negative logarithm of the marginal $\int_{\boldsymbol{\theta}_{est} } p({\bf y}_{i}, \boldsymbol{\theta}_{est} | {\bf x}_{i}, {\bf A} ) d\boldsymbol{\theta}_{est}$, we consider minimizing the negative logarithm of its maximum point-wise estimate, 
\textit{i.e.}, $\int_{\boldsymbol{\theta}_{est} } p({\bf y}_{i}, \boldsymbol{\theta}_{est} | {\bf x}_{i}, {\bf A} ) d \boldsymbol{\theta}_{est} \leq D p({\bf y}_{i}, \boldsymbol{\theta} | {\bf x}_{i}, {\bf A} )$, where we assume that $\boldsymbol{\theta}$ are the parameters for which $p({\bf y}_{i}, \boldsymbol{\theta}_{est} | {\bf x}_{i}, {\bf A} )$ has the maximum value and $D$ is a constant. Furthermore, 
we use the proportional relation \eqref{intro:joint:model.0.0.1} and by disregarding the partitioning function related to the prior \eqref{eq:Model:discriminate:parameters:model}, 
we end up with the following problem formulation:
\vspace{-.1in}
\begin{equation}
\begin{aligned}
&\{ \hat{\bf Y}, \hat{\bf A}, \hat{ \boldsymbol{\theta} } \} \! = \\
&\arg \min_{{\bf Y}, {\bf A}, \boldsymbol{\theta}}  \sum_{i=1}^{CK} \left[ \overbrace{\frac{1}{2}\Vert {\bf A}{\bf x}_i-{\bf y}_i \Vert_2^2 \! + \lambda_2 u_a( {\bf A}{\bf x}_i, \boldsymbol{\theta} )}^{-\log p({\bf x}_{i}|{\bf y}_{i}, \boldsymbol{\theta}, {\bf A})}   +  \right. \\
&\left. \overbrace{\lambda_0f_c( {\bf y}_i, \boldsymbol{\theta} )   + \lambda_1 \Vert  {\bf y}_i \Vert_1     +  \lambda_E  u_p(\boldsymbol{\theta} )}^{- \log p({\bf y}_{i}, \boldsymbol{\theta})} \right] + \overbrace{f_d({\bf A})}^{-\log p({\bf A})},
\end{aligned}
\label{intro:transform:model:general:pf}
\end{equation}
where 
$\{2, \lambda_0, \lambda_1, \lambda_2,\lambda_E \}$ are parameters inversely proportional to $\{\beta_0, \beta_d, \beta_1, \beta_a, \beta_E\}$. 

\subsection{The Learning Algorithm} 

Note that, solving \eqref{intro:transform:model:general:pf} jointly over $ {\bf A}$, $\boldsymbol{\theta}$ and ${\bf Y}$ is again challenging. 
Alternately, 
the solution of \eqref{intro:transform:model:general:pf} per any of the variables ${\bf A}, \boldsymbol{\theta}$ and ${\bf Y}$  
can be seen as an integrated 
marginal maximization (IMM) of 
$p({\bf Y}, {\bf A}|{\bf X})=p({\bf Y}|{\bf X}, {\bf A})p({\bf A}|{\bf X})$ that is approximated by $ \prod_{i=1}^{CK}p({\bf x}_{i}| {\bf y}_{i}, \boldsymbol{\theta},{\bf A})p({\bf y}_{i}, \boldsymbol{\theta}) p({\bf A}|{\bf x}_i)$, 
which is equivalent to: 
\vspace{-0.1in}
\begin{enumerate}
\item[1)] Approximately maximizing with $p({\bf x}_{i}|{\bf y}_{i},\boldsymbol{\theta}, {\bf A})$ and the prior $p(\boldsymbol{\theta},{\bf y}_{i})=p(\boldsymbol{\theta}|{\bf y}_{i})p({\bf y}_{i})$ over ${\bf y}_{i}$
\vspace{-0.051in}
\item[2)] Approximately maximizing with $\prod_{i=1}^{CK} p({\bf x}_{i}|{\bf y}_{i},\boldsymbol{\theta}, {\bf A})$ and the prior $p(\boldsymbol{\theta},{\bf y}_{i})=p({\bf y}_{i} | \boldsymbol{\theta})p(\boldsymbol{\theta})$ over $\boldsymbol{\theta}$
\vspace{-0.051in}
\item[3)] Approximately maximizing with $\prod_{i=1}^{CK} p({\bf x}_{i}|{\bf y}_{i},\boldsymbol{\theta}, {\bf A})$ and the prior $p( {\bf A})=p( {\bf A} |{\bf x}_i )$ over ${\bf A}$.
\end{enumerate}
\vspace{-0.1in}
In this sense, based on the IMM principle, we propose an iterative, alternating algorithm that
has three stages: (i) cluster and NT representation ${\bf y}_{i}$ assignment, 
(ii) discrimination parameters ${\boldsymbol{\theta}}$ update and (iii) linear map ${\bf A}$ update.

\label{Discriminative:Assignment:Step}

\vspace{-.1in}
{\flushleft \textbf{Stage 1: Cluster and NT Representation Assignment}} 
Given the data samples ${\bf X}$, the current estimate of ${\bf A}$, $\boldsymbol{\theta}$ and ${\bf Q}=$\footnote{Note that if ${\bf A}={\bf I}$ then ${\bf Q}={\bf X}$.}${\bf A}{\bf X}=[{\bf q}_1,...,{\bf q}_{CK}]$, the NT representations estimation problem is formulated as: 
\begin{equation}
\begin{aligned}
\! \! \! \! \! \! &\hat{{\bf Y}}=\\
\! \! \! \! \! \! &\arg \min_{{\bf Y}} \! \sum_{i=1}^{CK}\! \left[\frac{1}{2}\Vert {\bf q}_i-{\bf y}_i \Vert_2^2 \! + \! \lambda_1\Vert {\bf y}_{i} \Vert_1  \!+ \!
 \lambda_0f_c({\bf y}_{i}, \boldsymbol{\theta}) \right],
\end{aligned}
\label{eq:Total:PROJ:PROBLEM}
\end{equation}
where  $\{ \lambda_0, \lambda_1\}$ are inversely proportional to the scaling parameters $\{\beta_0, \beta_1\}$. 
Furthermore, given ${\bf q}_i$,
and $\boldsymbol{\theta}$, 
for any ${\bf y}_{i}$, 
\eqref{eq:Total:PROJ:PROBLEM} reduces to a
\textit{constrained projection problem}: 
\vspace{-.07in}
\begin{equation}
\begin{aligned}
\! \! \! \! \! \! \! \! \! \! \hat{{\bf y}}_i = 
\arg \min_{{\bf y}_i}&\left[ \frac{1}{2}\Vert {\bf q}_i-{\bf y}_i \Vert_2^2\!\!+\!\!\lambda_1{\bf 1}^T \vert {\bf y}_i \vert+ 
{f_c({\bf y}_i, \boldsymbol{\theta})} 
\right]= \\ 
\! \! \! \! \! \! \! \! \! \!  \arg \! \! \min_{ \icol{c_1 \in \mathcal{C}_d \\ c_2 \in \mathcal{C}_s  } }\! \! &\left\{ \min_{{\bf y}_i}\left[ \overbrace{\left( \frac{1}{2}\Vert {\bf q}_i-{\bf y}_i \Vert_2^2+\lambda_1{\bf 1}^T \vert {\bf y}_i \vert \right)}^{l_1(c_1,c_2)}+ \right. \right.\\
\! \! \! \! \! \! \! \! \! \!&\left. \left. \lambda_0\overbrace{ \left( \frac{\varrho({\bf y}_{i}, \boldsymbol{\tau}_{c_1})}{\varrho({\bf y}_{i}, \boldsymbol{\nu}_{c_2})}+\varsigma({\bf y}_{i}, \boldsymbol{\tau}_{c_1}) \right)}^{s_p(c_1,c_2)} \right] \right\},
\end{aligned}
\label{eq:Model:sparseFunc:reduction:problem:DP.1.0}
\end{equation}
where 
we derived the last expression by moving the minimization outwards from $f_c({\bf y}_i,\boldsymbol{\theta})\!\!=\!\!\min_{ \icol{c_1 \in \mathcal{C}_d \\  c_2 \in \mathcal{C}_s} }$ $\left[\right. \frac{\varrho({\bf y}_{i}, \boldsymbol{\tau}_{c_1})}{\varrho({\bf y}_{i}, \boldsymbol{\nu}_{c_2})}+\varsigma({\bf y}_{i}, \boldsymbol{\tau}_{c_1})\left. \right]$, while 
$\vert {\bf y}_i \vert$ is a vector whose elements are the absolute values of the elements in ${\bf y}_i$, thus ${\bf 1}^T\vert {\bf y} \vert=$ $\Vert {\bf y} \Vert_1$. 

In the following, we 
propose a solution to \eqref{eq:Model:sparseFunc:reduction:problem:DP.1.0}, which consists of two steps: (i) candidate NT representations estimation and (ii) 
cluster index and representation assignment. 


$-$\textit{\underline{ Candidate NT Representations Estimation} } Assuming that 
per each pair $\{ \boldsymbol{\tau}_{c_1}, \boldsymbol{\nu}_{c_2} \}$,$\{ c_1,c_2 \} \in \{ \mathcal{C}_d \times \mathcal{C}_s \}$, 
$\varrho(., \boldsymbol{\nu}_{c_2})\neq  0$, 
then the problem related to candidate NT representation estimation  considers only the cost $\left[ l_1(c_1,c_2)+\lambda_0s_P(c_1,c_2)\right]$ from \eqref{eq:Model:sparseFunc:reduction:problem:DP.1.0} and is defined as:
\vspace{-0.1in}
\begin{align}
\hspace{-1.45in}&{{\bf y}}|_{\{c_1,c_2\}} = \nonumber
\end{align}
\vspace{-0.2in}
\begin{equation}
\begin{aligned}
\! \! \! \! \! \! \! \! \! \!  \! \! \! \! \! \!& \arg \! \!  \min_{{\bf y}|_{\{c_1,c_2\}}}\left[   \overbrace{\left( \frac{1}{2}\Vert {\bf q}_i-{\bf y}|_{\{c_1,c_2\}} \Vert_2^2+\lambda_1{\bf 1}^T \vert {\bf y}|_{\{c_1,c_2\}} \vert \right) }^{l_1(c_1,c_2)}  + \right. 
\label{eq:Model:sparseFunc:reduction:problem:DP.1.0.single}
\end{aligned}
\end{equation}
\vspace{-0.3in}
\begin{align}
\! \! \! \! \!& \left. \lambda_0 \overbrace{\left( \frac{\varrho({\bf y}|_{\{  c_1, c_2  \}}, \boldsymbol{\tau}_{c_1})}{\varrho({\bf y}|_{\{c_1,c_2\}}, \boldsymbol{\nu}_{c_2})}+\varsigma({\bf y}|_{\{c_1,c_2\}}, \boldsymbol{\tau}_{c_1}) \right)}^{s_P(c_1,c_2)} \right]. \nonumber
\end{align}
\vspace{-0.in}
The closed form solution to \eqref{eq:Model:sparseFunc:reduction:problem:DP.1.0.single} is:
\vspace{-0.0in}
\begin{equation}
\begin{aligned}
\! \! \! \! \! \!{\bf y} |_{\{{c_1}, {c_2} \}}={\rm sign}({\bf q}_i) \odot \max \left( \vert {\bf q}_i\vert-{\bf t}_{c_1,c_2}, {\bf 0} \right)\oslash {\bf k}_{c_1},
\end{aligned}
\label{eq:Model:sparseFunc:reduction:solution:DP}
\end{equation}

\vspace{-0.1in}

{\flushleft where } 
$\odot$ and $\oslash$ are Hadamard product and division, while: 
\begin{align}
{\bf t}_{c_1, c_2}&=\lambda_0 (\frac{e_{c_1, c_2}}{h^2_{c_1, c_2}}{\bf v}_{c_2}+\frac{1}{h_{c_1, c_2}}{\bf g}_{c_1})-\lambda_1{\bf 1},\text{ ${}$}\text{ ${}$}\text{ ${}$} \nonumber \\
{\bf k}_{c_1}&=\left({\bf 1}+2\lambda_0\boldsymbol{\tau}_{c_1}\odot \boldsymbol{\tau}_{c_1} \right),\nonumber \\
{\bf g}_{c_1}&={\rm sign}({\bf q}_i^+)\odot \boldsymbol{\tau}_{c_1}^{+} +{\rm sign}(-{\bf q}_i^-)\odot \boldsymbol{\tau}_{c_1}^{-},\label{eq:Model:sparseFunc:reduction:solution:DP:pameters:0}\\
{\bf v}_{c_2}&={\rm sign}({\bf q}_i^+)\odot \boldsymbol{\nu}_{c_2}^{+}  +{\rm sign}(-{\bf q}_i^-) \odot \boldsymbol{\nu}_{c_2}^{-}. \nonumber
\end{align}

The variable $e_{c_1,c_2}$ is defined as:
\begin{equation}
\begin{aligned}
e_{c_1,c_2}&=\frac{ (h_{c_1,c_2}+c_s)^2 }{ h_{c_1,c_2}^2+\lambda_0{\bf g}_{c_1}^T{\bf v}_{k,c_2}  }\left({\bf g}_{c_1}^T\vert {\bf q}_{k,i}\vert-\right. \\
\lambda_0 &\left. \frac{1}{ h_{c_1,c_2}+c_s}{\bf g}_{c_1}^T{\bf g}_{k,c_1}-\lambda_1{\bf g}_{c_1}^T{\bf l}_{k,c_1}-c_s\right),
\end{aligned}
\label{eq:Model:sparseFunc:reduction:solution:DP:pameters:1}
\end{equation}
where $c_s=0$, ${\bf v}_{k,c_2}={\bf v}_{c_2}\oslash{\bf k}_{c_1}, {\bf g}_{k,c_1}={\bf g}_{c_1}\oslash{\bf k}_{c_1}, \vert{\bf q}_{k,i}\vert=\vert {\bf q}_i \vert \oslash{\bf k}_{c_1}$ and $h_{c_1,c_2}$ is a solution to a quartic polynomial 
(\textit{Appendix} A).

\textit{$-$ \underline{Assignment} }
\label{DiscriminativeAssignment:Clustering}
This step consists of two parts.

\fbox{\begin{minipage}{4em}
{\textit{ ${ }$ Part $1$}}
\end{minipage}}
Given all ${\bf y}|_{\{{c_1}, {c_2} \}}$, $  \{c_1, c_2 \} \in \mathcal{C}_d \times \mathcal{C}_s$, the first part evaluates a score related to $f_c({\bf y}_{i},\boldsymbol{\theta})$ 
as follows: 
\vspace{-0.05in}
\begin{align}
s_P(c_1,c_2)=&\frac{\varrho({\bf y}|_{\{{c_1}, {c_2} \}}, \boldsymbol{\tau}_{c_1})}{\varrho({\bf y}|_{\{{c_1}, {c_2} \}}, \boldsymbol{\nu}_{c_2})}+\varsigma({\bf y}|_{\{ {c_1}, {c_2} \}}, \boldsymbol{\tau}_{c_1}).\label{eq:Model:sparseFunc:assiment:score:DP} 
\end{align}
\vspace{-0.4in}
{\flushleft \text{}} 

\fbox{\begin{minipage}{4em}
{\textit{ ${ }$ Part $2$}}
\end{minipage}} In the second part, we assume that the costs $l_1(c_1,c_2)$ in the respective subproblems  \eqref{eq:Model:sparseFunc:reduction:problem:DP.1.0.single} across all of the estimated ${\bf y}|_{\{c_1,c_2\}}$ are approximatively equal, \textit{i.e.}: 
\begin{equation}
\begin{aligned}
& \frac{1}{2}\Vert {\bf q}_i-{\bf y}|_{\{1,1\}} \Vert_2^2+\lambda_1{\bf 1}^T \vert {\bf y}|_{\{1,1 \}} \vert  \simeq. ... \simeq \\  
&  \frac{1}{2}\Vert {\bf q}_i-{\bf y}|_{\{C_d,C_s\}} \Vert_2^2+\lambda_1{\bf 1}^T \vert {\bf y}|_{\{C_d,C_s\}} \vert,
\label{eq:Model:sparseFunc:assiment:min:evluation:0.0.1}
\end{aligned}
\end{equation}
which is a reasonable assumption when the sparsity level $\lambda_1$ is same for all ${\bf y}|_{ \{c_1,c_2\} }, c_1 \in \mathcal{C}_d, c_2 \in \mathcal{C}_s$.  
Therefore, we disregard $l_1(c_1,c_2)$  and based on the score \eqref{eq:Model:sparseFunc:assiment:score:DP}, we assign the cluster index and the NT representation ${\bf y}_{i}$ as follows:  
\begin{align}
\hat{{\bf y}}_{i}=&{\bf y}|_{\{j_1(i), j_2(i) \}}, \label{eq:Model:sparseFunc:assiment:min:evluation} \\
\{\overbrace{\hat{c}_1}^{j_1(i)}, \overbrace{\hat{c}_2}^{j_2(i)} \}=&\arg \min_{\icol{c_1 \in \mathcal{C}_d}} \min_{c_2 \in \mathcal{C}_s}  s_P(c_1,c_2), \nonumber 
\end{align}
\vspace{-0.28in}
{\flushleft } 
where the evaluation w.r.t. 
$f_c({\bf y}_{i},\boldsymbol{\theta})$ 
reduces to computing a minimum score over $s_P(.)$ 
as in \eqref{eq:Model:sparseFunc:assiment:min:evluation}. 

\vspace{-0.1in}
{\flushleft \textbf{Stage 2: Parameters $\boldsymbol{\theta}$ Update}} 
Given the estimated NT representations ${\bf Y}=[{\bf y}_{1},..,{\bf y}_{CM}]$, the linear map ${\bf A}$ and ${\bf Q}={\bf A}{\bf X}$, 
the problem related to update of the parameters $\boldsymbol{ \theta}$  
reduces to the following form:
\vspace{-0.in}
\begin{equation}
\begin{aligned}
\! \! \! \! \! \! \! \! \! \! \! \hat{\boldsymbol{\theta}}= \arg \min_{ \boldsymbol{\theta}  } \sum_{i=1}^{CK}\left[ u_a( {\bf q}_{i}, \boldsymbol{\theta} ) \! \! + \! \! \lambda_0f_c({\bf y}_{i}, \boldsymbol{\theta}) \right] \! \! + \! \!  \lambda_Eu_{p}(\boldsymbol{\theta}), \label{eq:Proposed:Thetha:parameters:Update}
\end{aligned}
\end{equation} 
\vspace{-0.26in}

{\flushleft where} 
$\lambda_E$ is inversely proportional to  $\beta_E$ and $u_{p}(\boldsymbol{\theta})$ is the measure described in section \ref{prir:measure}.  Note that in the cluster and NT representation assignment step 
(\textbf{Stage 1}, part 2 of our algorithm), for each ${\bf y}_{i}$ the corresponding $\boldsymbol{\tau}_{c_1}$ and $\boldsymbol{\nu}_{c_2}$ are known $(A_{ss}):\{ {\bf y}_i, \{ {\bf y}|_{\{j_1(i),j_2(i)\}}, \boldsymbol{\tau}_{j_1(i)}, \boldsymbol{\nu}_{j_2(i)}\} \}$. 
Therefore, at this stage, we do not evaluate the terms 
$f_c({\bf y}_{i}, \boldsymbol{\theta})$ w.r.t. $\boldsymbol{\theta}$. Instead, we use the already evaluated scores based on the assignment w.r.t. ${\bf y}_i$. 

In the following, we present the problems related to update of the parameters $\boldsymbol{\tau}_{c_1}$ and $\boldsymbol{\nu}_{c_2}$ and comment on the solutions, which represent a slight extension to the previous one. 

{\textit{$-$ \underline{ Update Per Single $\boldsymbol{\tau}_{c_1}$ } } }
Given ${\bf Q}={\bf A}{\bf X}$, ${\bf Y}$, ${\boldsymbol{\theta}_{\setminus  c_1 }}$ 
and using $( A_{ss} )$, 
problem \eqref{eq:Proposed:Thetha:parameters:Update}, 
per ${\boldsymbol{\tau}_{c_1}}$ reduces to:
\vspace{-0.05in}
\begin{equation}
\begin{aligned}
\! \! \! \! \! \! \! \! & \hat{\boldsymbol{\tau}}_{c_1}=\arg\min_{ {\boldsymbol{\tau}_{c_1}} }  \sum_{\icol{ \forall i: \\ j_1(i)==c_1 }} \left[ \frac{1}{2} \Vert {\bf q}_{i}-\boldsymbol{\tau}_{c_1}-\boldsymbol{\nu}_{j_{2}(i)} \Vert_2^2+\right. \\
\! \! \! \! \! \! \! \!  & \left. \lambda_0  \left( \frac{\varrho({\bf y}_{i}, \boldsymbol{\tau}_{c_1})}{\varrho({\bf y}_{i}, \boldsymbol{\nu}_{j_{2}(i)})}+\varsigma({\bf y}_{i}, \boldsymbol{\tau}_{c_1}) \right) \right]+  \lambda_E f_c(\boldsymbol{\tau}_{c_1}, \boldsymbol{\theta}_{\setminus c_1} ).
\end{aligned}
\label{eq:Proposed:Dis:Discriminative:Parameters:all:nu:DD}
\end{equation} 
The solution for \eqref{eq:Proposed:Dis:Discriminative:Parameters:all:nu:DD} 
is similar to the solution given by \eqref{eq:Model:sparseFunc:reduction:solution:DP},  \eqref{eq:Model:sparseFunc:assiment:score:DP} and \eqref{eq:Model:sparseFunc:assiment:min:evluation}. 
That is, compared to  \eqref{eq:Model:sparseFunc:reduction:solution:DP}, in the solution of  \eqref{eq:Proposed:Dis:Discriminative:Parameters:all:nu:DD}, 
the part related to candidate NT estimation, both the respective thresholding and normalization vectors have additional terms 
 (we give the exact expression and proof in \textit{Appendix} B.1).

{\textit{$-$ \underline{ Update Per Single $\boldsymbol{\nu}_{c_2}$ } } }
Given ${\bf Q}={\bf A}{\bf X}$, ${\bf Y}$, $\boldsymbol{\theta}_{\setminus c_2 }$ 
and using $( A_{ss} )$,  
problem \eqref{eq:Proposed:Thetha:parameters:Update}, 
per ${\boldsymbol{\nu}_{c_2}}$ reduces to:
\vspace{-0.05in}
\begin{equation}
\begin{aligned}
\!  \! \! \! \! \! \! \! \!  \hat{\boldsymbol{\nu}}_{c_2}=\arg \min_{ {\boldsymbol{\nu}_{c_2}} } & \sum_{\icol{\forall i: \\ j_2(i)==c_2} }  \left[ \frac{1}{2}\Vert {\bf q}_{i}- \boldsymbol{\tau}_{j_1(i)}-{\boldsymbol{\nu}_{c_2}} \Vert_2^2+
\right.  \\
\! \! \! \! \! \! \! \! & \left.  \lambda_0 \frac{\varrho({\bf y}_{i}, \boldsymbol{\tau}_{j_1(i)})}{\varrho({\bf y}_{i}, \boldsymbol{\nu}_{c_2})} \right]+\lambda_E f_c(\boldsymbol{\nu}_{c_2}, \boldsymbol{\theta}_{\setminus c_2} ).
\end{aligned}
\label{eq:Proposed:Dis:Discriminative:Parameters:all:nu:DP}
\end{equation} 
\vspace{-0.24in}

{\flushleft In this update}, \eqref{eq:Proposed:Dis:Discriminative:Parameters:all:nu:DP} 
is solved 
iteratively, where per each iteration the solution  for the candidate NT representation is similar 
to the solution for \eqref{eq:Model:sparseFunc:reduction:solution:DP}, but  
$\boldsymbol{\nu}_{c_2}$ is estimated using different 
thresholding and normalization vectors (for the exact expression and proof please see \textit{Appendix} B.2).


\vspace{-0.1in}
{\flushleft \textbf{Stage 3: Linear Map ${\bf A}$ Update} }
Given the data samples ${\bf X}$, the corresponding transform representations ${\bf Y}$ and the discrimination parameters $\boldsymbol{\theta}$, 
the problem related to the estimation of the linear map ${\bf A}$, reduces to:
\begin{equation}
\begin{aligned}
\hat{\bf A}=\arg  &\min_{{\bf A}}\frac{1}{2}\Vert {\bf A}{\bf X}-{\bf Y}_T \Vert_F^2+ 
\frac{\lambda_2}{2}\Vert {\bf A} \Vert_{F}^2+\\
&\frac{\lambda_3}{2}\Vert {\bf A}{\bf A}^T-{\bf I} \Vert_F^2-\lambda_4\log \vert \det {\bf A}^T{\bf A} \vert,
\end{aligned}
\label{eq:Proposed:lin:mapEst}
\end{equation} 
where $\Vert {\bf A}{\bf X}-{\bf Y}_T \Vert_F^2=\frac{1}{2}\Vert {\bf A}{\bf X}-{\bf Y} \Vert_F^2+\frac{1}{2}\sum_{i=1}^{CK}\Vert {\bf A}{\bf x}_{i}- \boldsymbol{\tau}_{j_1(i)}-{\boldsymbol{\nu}_{j_2(i)}} \Vert_2^2$ and ${\bf Y}_T$ is expressed as:
\begin{equation}
\begin{aligned}
{\bf Y}_T =&\left[ {\bf y}_1+\boldsymbol{\tau}_{j_1(1)}+\boldsymbol{\nu}_{j_2(1)},...,\right.\\
&\left. {\bf y}_{CK}+\boldsymbol{\tau}_{j_1(CK)}+\boldsymbol{\nu}_{j_2(CK)} \right],
\end{aligned}
\label{eq:Proposed:lin:mapEst.YT}
\end{equation} 
while $\boldsymbol{\tau}_{j_1(i)}$ and $\boldsymbol{\nu}_{j_2(i)}$ denote the corresponding $\boldsymbol{\tau}_{c_1}$ and $\boldsymbol{\nu}_{c_2}$ that appear in the NT, which is used to estimate ${\bf y}_i$, $\forall i \in \{1 , ..., CK\}$ and the parameters $\{ \lambda_2, \lambda_3, \lambda_4 \}$ are inversely proportional to the scaling parameters $\{\beta_3, \beta_4, \beta_5 \}$. We solve \eqref{eq:Proposed:lin:mapEst} using an approximate closed form solution as proposed in \cite{Kostadinov2018:EUVIP}. 

We point out that when 
${\bf A}={\bf I}$, then this stage is omitted and our learning algorithm reduces to alternating between cluster and representation ${\bf Y}$ assignment and update of $\boldsymbol{\theta}$.



\vspace{-0.1in}
\section{Approach Evaluation}
This section 
evaluates the advantages and the potential of the proposed algorithm and compares its  clustering performance to the state-of-the-art methods.




\subsection{Data, Setup and Measures}

{\flushleft \textbf{Data Sets} } 
The used data sets are E-YALE-B \cite{extYaleB:Georghiades01}, AR \cite{ARDATABSE:Martinez1998}, ORL \cite{Samaria94parameterisationof} and COIL \cite{COIL20:Nene96}. All the images from the respective datasets were downscaled to resolutions $21\times21$, $32\times28$, $24\times24$ and $20\times25$,  
respectively, and are normalized to unit variance. 

\vspace{-0.1in}
{\flushleft \textbf{Algorithm and Clustering Set Up} } The used setup is described in the following text.

\textit{$-$ On-Line Version} An on-line variant is used for the update of ${\bf A}$ w.r.t. a subset of the available training set. It has the following form ${\bf A}^{t+1}={\bf A}^{t}-\rho({\bf A}^{t}-\hat{{\bf A}} )$ where $\hat{{\bf A}}$ and ${\bf A}^t$ are the the solutions in the transform update step at iterations $t+1$ and  $t$, which is equivalent to having the additional constraint ${\Vert {\bf A}^t -\hat{{\bf A}}\Vert_F^2}$ in the related problem. 
The used batch size is equal to $87\%, 85\%, 90\%$ and $ 87\%$ 
of the total amount of the available training data from the respective datasets E-YALE-B, AR, ORL and COIL. 

\textit{$-$ Clustering Setup, Cluster Index and NT Estimation } We assume that the number of clusters $C$ per database is known. 
We set  the number of parameters that are related to dissimilarity $\boldsymbol{\tau}_{c_1}, c_1 \in \{1,...,C_d\}$ to be close to the number of actual clusters $C$, \textit{i.e.}, $C_d=C$
and we set the number of parameters $\boldsymbol{\nu}_{c_2}, c_2 \in \{1,...,C_s\}$ related to similarity to be small, \textit{i.e.}, $C_s$ is small. 
The cluster index $c$ 
and the NT are estimated based on the minimum score of the discriminative functional measure as explained in Section \ref{DiscriminativeAssignment:Clustering}. 
As an evaluation metric for the clustering performance 
we use the cluster accuracy (CA) and the normalized mutual information (NMI) \cite{Cai:2011:GRN}. 

\textit{$-$ Algorithm Set-up} The parameters $\lambda_0=\lambda_1=0.03$, $\lambda_E=0.001$, $\lambda_2=\lambda_3=\lambda_4=16$, the transform dimension is $M=2100$. 
The algorithm is initialized with ${\bf A}$ and $\boldsymbol{\theta}$ having i.i.d. Gaussian (zero mean, unit variance) entries and is terminated after the $100$th iteration. The results are obtained as the average of $5$ runs. 



\vspace{ -0.1in}
\begin{table}[t]
\vspace{ -0.06in}
\centering
\begin{minipage}[b]{1\linewidth}
\centering
\begin{tabular}{             l@{\hspace{-14.5pt}}c@{\hspace{1pt}}
				  c@{\hspace{1pt}}|c@{\hspace{1pt}}|c@{\hspace{1pt}}|
                  c@{\hspace{1pt}}|c@{\hspace{1pt}}|c@{\hspace{1pt}}|
                  c@{\hspace{1pt}}|c@{\hspace{1pt}}|c@{\hspace{1pt}}|
                  c@{\hspace{1pt}}|c@{\hspace{1pt}}|c@{\hspace{1pt}}}

&&\multicolumn{3}{c|}{COIL}&\multicolumn{3}{c|}{ORL}&\multicolumn{3}{c|}{E-YALE-B}&\multicolumn{3}{c}{AR} \\
\cline{3-14} 
 && ${\kappa}_{n}$ & $\mu$ & $t$  & ${\kappa}_{n}$ & $\mu$ & $t$  & ${\kappa}_{n}$ & $\mu$ & $t$& ${\kappa}_{n}$ & $\mu$ & $t$  \\
\hline
 & \text{ ${}$}   &  16& .2e-5& 46&21& .3e-5&  48& 31&  .1e-5& 51&  28& .3e-5&  69
\end{tabular}
\end{minipage}
\vspace{-.0in}
\caption{The computational efficiency per iteration $t[sec]$ for the proposed algorithm, the conditioning number ${\kappa}_{n}({\bf A})=\frac{\sigma_{max}}{\sigma_{min}}$ and the expected mutual coherence $\mu({\bf A})$ for the liner map ${\bf A}$. }
\label{cP:Part:1}
\vspace{-.049in}
\end{table}

\subsection{Numerical Experiments}

{\flushleft \textbf{Summary}} Our experiments consist of three parts. 

\begin{table}[t]
\centering
\begin{minipage}[b]{1\linewidth}
\centering
\begin{tabular}{r|c|c|c|c}
 & COIL & ORL & E-YALE-B & AR   \\
\hline
{CA $\%$} \text{ ${}$}  & $89.2$ & $75.4$ & $96.8$ & $94.8$ \\
  \hline
NMI $\%$ \text{ ${}$}  \text{$ $}& $91.2$ & $84.1$ & $95.3$ & $94.1$
\end{tabular}
\end{minipage}
\vspace{-.1in}
\caption{The clustering performance over the databases COIL, ORL, E-YALE-B and AR evaluated using the Cluster Accuracy (CA) and the Normalized Mutual Information (NMI) metrics.}
\label{cP:Part:2}
\vspace{-.0in}
\end{table}

\textit{$-$ NT Properties} In the first series of the experiments, we investigate the properties of the proposed algorithm. 
We measure the run time $t$ of the proposed algorithm, the conditioning number ${\kappa}_{n}({\bf A})=\frac{\sigma_{max}}{\sigma_{min}}$ ($\sigma_{min}$ and $\sigma_{max}$ are the smallest and the largest singular values of ${\bf A}$, respectively) and the expected mutual coherence $\mu({\bf A})$ as in \cite{Kostadinov2018:EUVIP} of the shared linear map ${\bf A}$ in the learned NTs. 

\textit{$-$ Clustering and k-NN Classification Performance} In the second part, we measure the performance across all databases and report the CA and NMI. We also split every databases on training and test set and learn NTs with the proposed algorithm on the training set. We use the learned NTs to assign a representation for the test data and then preform a k-NN \cite{Cover:2006:EIT:1146355} search using the test NT representation on the training NT representation.

\vspace{-0.00in}

\textit{$-$ Proposed Method vs State-Of-The-Art} 
This part compares the proposed method w.r.t. results reported by five state-of-the-art methods, including: 
GSC \cite{zheng2011graph},  NSLRR \cite{yin2016laplacian},  SDRAM \cite{guo2015robust} and RGRSC\cite{kodirov2016learning}. 


\begin{table}
\vspace{-.in}
\begin{minipage}[b]{0.49\linewidth}
\centering
\begin{tabular}{@{\hspace{0pt}}l@{\hspace{4pt}}|c@{\hspace{4pt}}|c@{\hspace{4pt}}|c@{\hspace{4pt}}}
&\multicolumn{3}{c}{ CA $\%$ } \\
\cline{2-4} 
\text{$ $} & COIL & ORL & E-YALE-B  \\
\cline{1-4} 
CASS \cite{lu2013correlation} & 59.1 &68.8 &81.9 \\
GSC \cite{zheng2011graph} & 80.9 &61.5 &74.2 \\
NSLRR \cite{yin2016laplacian} & 62.8 &55.3 &/ \\
SDRAM \cite{guo2015robust} & 86.3 &70.6 &92.3 \\
RGRSC \cite{kodirov2016learning} &88.1 &$\bf  76.3$ &95.2 \\
$(*)$& $\bf 89.2$ & $75.4$ & $\bf 96.8$
\end{tabular}
\end{minipage}
\caption{A comparative results between state-of-the-art \cite{lu2013correlation}, \cite{zheng2011graph}, \cite{yin2016laplacian}, \cite{guo2015robust} and \cite{kodirov2016learning} and the proposed method $(*)$.}
\label{cP:Part:3}
\vspace{-.12in}
\end{table}

\begin{table}
\vspace{-.0in}
\begin{minipage}[b]{0.49\linewidth}
\centering
\begin{tabular}{@{\hspace{0pt}}l@{\hspace{4pt}}|c@{\hspace{4pt}}|c@{\hspace{4pt}}|c@{\hspace{4pt}}}
&\multicolumn{3}{c}{ NMI $\%$ } \\
\cline{2-4} 
\text{$ $} & COIL & ORL & E-YALE-B  \\
\cline{1-4} 
CASS \cite{lu2013correlation} & 64.1 &78.1 &78.1 \\
GSC \cite{zheng2011graph} & 87.5 &76.2 &75.0 \\
NSLRR \cite{yin2016laplacian} & 75.6 &74.5 &/ \\
SDRAM \cite{guo2015robust} & 89.1 &80.2 &89.1 \\
RGRSC \cite{kodirov2016learning} & 89.3 &$\bf  86.1$ &94.2 \\
$(*)$& $\bf 91.2$ & $84.1$ & $\bf 95.3$
\end{tabular}
\end{minipage}
\caption{A comparative results between state-of-the-art \cite{lu2013correlation}, \cite{zheng2011graph}, \cite{yin2016laplacian}, \cite{guo2015robust} and \cite{kodirov2016learning}, and the proposed method $(*)$.}
\label{cP:Part:4}
\vspace{-.1in}
\end{table}


\begin{table}
\vspace{-.0in}
\centering
\begin{minipage}[b]{1\linewidth}
\centering
\begin{tabular}{l|c|c|c|c}
 & COIL & ORL & E-YALE-B & AR   \\
\hline
{acc. NT } \text{ ${}$}  & $97.1$ & $96.9$ & $96.8$ & $96.0$ \\
\hline
{acc. OD} \text{ ${}$}  & $94.0$ & $94.5$ & $93.4$ & $91.6$
\end{tabular}
\end{minipage}
\vspace{-.15in}
\caption{The k-NN accuracy results using assigned NT representations and original data (OD) representation.}
\label{cP:Part:Recognition}
\vspace{-.19in}
\end{table}

\vspace{-0.15in}
{\flushleft \textbf{Evaluation Results}} We show the results in Tables \ref{cP:Part:1}, \ref{cP:Part:2}, \ref{cP:Part:3}, \ref{cP:Part:4} and \ref{cP:Part:Recognition}. 

{$-$ \textit{NT Properties}} 
As shown in Table \ref{cP:Part:1}, the learned NTs for all the data sets have relatively low computational time per iteration. 
All linear maps in the NTs have good conditioning numbers and low expected coherence. 

{$-$ \textit{Clustering Performance}} The results of the clustering performance over the databases E-YALE-B \cite{extYaleB:Georghiades01}, AR \cite{ARDATABSE:Martinez1998}, ORL \cite{Samaria94parameterisationof} and COIL \cite{COIL20:Nene96} 
are shown in Table \ref{cP:Part:2}. We see that both the CA and the NMI measures have high values.  The highest performance is reported on the E-YALE-B \cite{extYaleB:Georghiades01} databases where the CA and NMI are $96.8 \%$ and  $95.3 \%$, respectively.

{$-$ \textit{Proposed vs State-Of-The-Art Clustering}} The results are shown on Tables \ref{cP:Part:3} and \ref{cP:Part:4}.  As we see the proposed algorithm outperforms the state-of-the-art methods CASS \cite{lu2013correlation},  GSC \cite{zheng2011graph},  NSLRR \cite{yin2016laplacian},  SDRAM \cite{guo2015robust} and RGRSC\cite{kodirov2016learning}. The highest gain in CA and NMI w.r.t. the state-of-the-art is $1.6 \%$ and  $1.9 \%$, respectively, that is achieved on the E-YALE-B \cite{extYaleB:Georghiades01} and the COIL \cite{COIL20:Nene96} databases, respectively.

{$-$ \textit{k-NN Classification Performance}} The results of the k-NN performance on all databases is shown in Table  \ref{cP:Part:Recognition}. As a baseline we use k-NN on the original data and report improvements of $3.1 \%$, $2.4 \%$, $3.3 \%$ and $4.4 \%$ over the baseline results for the respective databases.

\vspace{-0.1in}
\section{Conclusion}
\vspace{-0.05in}

In this paper, we modeled assignment based NT with priors. 
A novel clustering concept was introduced where we (i) jointly learn the NTs with priors and 
(ii) assign the cluster and the NT representation based on maximum likelihood over functional measure. 
Given the observed data, an empirical approximation to the maximum likelihood of the model gives the corresponding problem formulation. We proposed an efficient solution for learning the model parameters by a low complexity iterative alternating algorithm. 

The proposed algorithm was evaluated on publicly available databases. The preliminary results showed promising performance. 
In a clustering regime w.r.t. the used CA and NMI measures, the algorithm gives improvements compared to the state-of-the-art methods. In unsupervised k-NN classification regime, 
it demonstrated high classification accuracy.

\bibliography{aistat2019}
\bibliographystyle{plainnat}
\end{document}